\documentclass{article}
\usepackage{ijcai19}

\usepackage{times}
\usepackage{soul}
\usepackage{url}
\usepackage[utf8]{inputenc}
\usepackage[small]{caption}
\usepackage{graphicx}
\usepackage{amsmath}
\usepackage{mathtools}
\usepackage{booktabs}
\usepackage{verbatim}
\usepackage[colorinlistoftodos]{todonotes}

\usepackage{natbib}
\usepackage{algorithm}
\usepackage[algo2e]{algorithm2e}

\usepackage{CommentingStyle}

\usepackage{lipsum}

\newif\ifcomments

\commentstrue

\ifcomments
\newcommand{\comments}[1]{#1}
\else
\newcommand{\comments}[1]{}
\fi

\title{Towards Empathic Deep Q-Learning}

\author{
Bart Bussmann$^1$
\and
Jacqueline Heinerman$^2$\And
Joel Lehman$^3$
\affiliations
$^1$University of Amsterdam, The Netherlands\\
$^2$VU University Amsterdam, The Netherlands\\
$^3$Uber AI Labs, San Francisco, United States
\emails
bart.bussmann@student.uva.nl, jacqueline@heinerman.nl,
joel.lehman@uber.com
}

\begin{document}

\maketitle

\begin{abstract}
As reinforcement learning (RL) scales to solve increasingly complex tasks, interest continues to grow in the fields of AI safety and machine ethics.
As a contribution to these fields, this paper introduces an extension to Deep Q-Networks (DQNs), called \textit{Empathic DQN}, that is loosely inspired both by empathy and the golden rule (``Do unto others as you would have them do unto you''). Empathic DQN aims to help mitigate \emph{negative side effects} to other agents resulting from myopic goal-directed behavior. We assume a setting where a learning agent coexists with other independent agents (who receive unknown rewards), where some types of reward (e.g.\ negative rewards from physical harm) may generalize across agents.
Empathic DQN combines the typical (self-centered) value with the estimated value of other agents, by imagining (by its own standards) the value of it being in the other's situation (by considering constructed states where both agents are swapped). 
Proof-of-concept results in two gridworld environments highlight the approach's potential to decrease collateral harms. While extending Empathic DQN to complex environments is non-trivial, we believe that this first step highlights the potential of bridge-work between machine ethics and RL to contribute useful priors for norm-abiding RL agents.
\end{abstract}

\section{Introduction}

Historically, reinforcement learning (RL; \citep{sutton1998introduction}) research has largely focused on solving clearly-specified benchmark tasks. For example, 
the ubiquitous Markov decision process (MDP) framework cleaves the world into four well-defined parts (states, actions, state-action transitions, and rewards), and most RL algorithms and benchmarks 
leverage or reify the assumptions of this formalism, e.g.\ that a singular, fixed, and correct reward function exists, and is given. While there has been much exciting progress in learning to solve
complex well-specified tasks (e.g.\ super-human performance in go \citep{silver2016mastering} and Atari \citep{mnih2015human}), there is also increasing recognition that common 
RL formalisms are often meaningfully imperfect 
\citep{hadfield2017inverse,lehman2018surprising}, and that there remains much to understand
about safely applying RL to solve real-world tasks \citep{amodei2016concrete}. 


As a result of this growing awareness, there has been increasing interest in the field of AI safety \citep{amodei2016concrete,everitt2018agi}, which is broadly concerned with creating AI agents that do what is \emph{intended} for them to do, and which often
entails questioning and extending common formalisms \citep{hadfield2017inverse,hadfield2016cooperative,demski2019embedded}. One overarching theme in AI safety 
is how to learn or provide correct incentives to an agent. \citet{amodei2016concrete} distinguishes different failure modes in specifying reward functions, which include \emph{reward hacking}, wherein an
agent learns how to optimize the reward function in an unexpected and unintended way that does not satisfy the underlying goal, and \emph{unintended side effects}, wherein an agent learns to achieve the desired goal, but
causes undesirable collateral harm (because the given reward function is incomplete, i.e.\ it does not include all of the background knowledge and context of the human reward designer). 

This paper focuses on the latter setting, i.e.\ assuming that the reward function incentivizes solving the task, but fails to anticipate some unintended harms. We assume that in real world settings, a physically-embodied RL agent (i.e.\ a controller for a robot) will often share space with other agents (e.g.\ humans, animals, and other trained computational agents), and it is challenging to design reward functions apriori that enumerate all the ways in which other agents can be negatively affected \citep{amodei2016concrete}. Promising current approaches include value learning from human preferences \citep{saunders2018trial,leike2018scalable,hadfield2016cooperative} and creating agents that attempt to minimize their impact on
the environment \citep{krakovna2018measuring,turner2019conservative}; however, value learning can be expensive for its need to include humans in the loop, and both directions remain technically and philosophically challenging. This paper introduces another tool that could complement such existing approaches, motivated by the concept of empathy. 

In particular, the insight motivating this paper is that humans often empathize with the situations of others, by generalizing from their own past experiences. 
For example, we can feel vicarious fear for someone who is walking a tight-rope, because we ourselves would be afraid in such a situation. Similarly, for some classes of reward signals (e.g.\ physical harm), it may be reasonable for 
embodied computational agents to generalize those rewards to other agents (i.e.\ to assume as a prior expectation that other agents might receive similar reward in similar situations). If a robot learns that a fall from heights is dangerous to itself, that insight could generalize to most other embodied agents. 

For humans, beyond granting us a capacity to understanding others, such empathy
also influences our behavior, e.g.\ by avoiding harming others while walking down the street; likewise, in some situations it may be useful if learning agents could also act out of empathy (e.g.\ to prevent physical harm to another agent resulting from otherwise blind goal-pursuit). While there are many
ways to instantiate algorithms that abide by social or ethical norms (as studied by the field of \emph{machine ethics} \citep{anderson2011machine,wallach2008moral}), here we take loose inspiration from one simple ethical norm, i.e.\ the golden rule.

The golden rule, often expressed as: ``Do unto others
as you would have them do unto you,'' is a principle that has
emerged in many ethical and religious contexts \citep{kng1993global}.
At heart, abiding by this rule entails projecting one's desires onto another agent,
and attempting to honor them. We formalize this idea as an extension
of Deep Q-Networks (DQNs; \citep{mnih2015human}), which we 
call \emph{Empathic DQN}. The main idea is to augment the value of a given
state with the value of constructed states simulating what the learning agent
would experience if its position were switched with another agent.
Such an approach can also be seen as learning to maximize an estimate of
the combined rewards of both agents, which embodies 
a utilitarian ethic.

The experiments in this paper apply Empathic DQN to two
gridworld domains, in which a learned agent pursues a goal in an
environment shared with other non-learned (i.e.\ fixed) agents. In
one environment, an agent can harm and be harmed by other agents;
and in another, an agent receives diminishing returns from hoarding
resources 
that also could benefit other agents. 
Results in these domains show that Empathic DQN can reduce negative side effects in both environments. 
While much work is needed before this algorithm would be
effectively applicable to more complicated environments,
we believe that this first step highlights the possibility
of bridgework between the field of machine ethics and
RL; in particular, for the purpose of instantiating useful 
priors for RL agents interacting
in environments shared with other agents.

\section{Background}

This section reviews machine ethics and AI safety, two fields studying how to encourage and ensure acceptable behavior in computational agents.

\subsection{Machine Ethics}

The field of machine ethics \citep{anderson2011machine,wallach2008moral}
studies how to design algorithms (including RL algorithms) capable of moral behavior. While morality
is often a contentious term, with no agreement among moral philosophers
(or religions) as to the nature of a ``correct'' ethics, from a 
pragmatic viewpoint, agents deployed in the real world will encounter
situations with ethical tradeoffs, and to be palatable their behavior will need to 
approximately satisfy certain societal and legal norms. Anticipating
and hard-coding acceptable behavior for all such trade-offs is likely
impossible. Therefore, just as humans take ethical stances in the real
world in the absence of universal ethical consensus, we may need the same pragmatic behavior from intelligent
machines.

Work in machine ethics often entails concretely embodying a particular moral
framework in code, and applying the
resulting agent in its appropriate domain. For example, 
\citet{winfield2014towards} implements a version of
Asmiov's first law of robotics (i.e.\ ``A robot may not injure a human being or, through inaction,
allow a human being to come to harm'') in a wheeled
robot that can intervene to stop another robot (in lieu of an actual human)
from harming itself. Interestingly, the implemented system bears a strong 
resemblance to model-based RL; such reinvention, and the strong possibility
that agents tackling complex tasks with ethical dimensions 
will likely be driven by machine learning (ML), 
suggests the potential benefit 
and need for increased cooperation between ML and machine
ethics, which is an additional motivation for our work.

Indeed, our work can be seen as a contribution to the intersection of
machine ethics and ML, in that
the process of empathy is an important contributor to morally-relevant
behavior in humans \citep{tangney2007moral}, and that to the 
authors' knowledge, there has not been previous work implementing
golden-rule-inspired architectures in RL.

\subsection{AI Safety}

A related but distinct field of study is AI safety \citep{amodei2016concrete,everitt2018agi},
which studies how AI agents can be implemented to avoid harmful accidents. Because
harmful accidents often have ethical valence, there is necessarily overlap between the two fields,
although technical research questions in AI safety may not be phrased in the language
of ethics or morality.

Our work most directly relates to the problem of negative side-effects, as described by \citet{amodei2016concrete}. In this problem the designer specifies an objective function that focuses on accomplishing a specific task (e.g.\ a robot should clean a room),
but fails to encompass all other aspects of the environment (e.g.\ the robot should not vacuum the cat); the result is an agent that is indifferent to whether it alters the environment in undesirable ways, e.g.\ causing harm to the cat. 
Most approaches to mitigating side-effects aim to generally minimize the impact the agent has on the environment through intelligent heuristics \citep{armstrong2017low,krakovna2018measuring,turner2019conservative}; we believe that other-agent-considering heuristics (like ours) are likely
complementary. Inverse reinforcement learning (IRL; \citep{abbeel2004apprenticeship}) aims to directly learn the rewards of other agents (which a learned agent could then take into account) and could also be meaningfully combined with our approach (e.g.\ Empathic DQN could serve as a prior
when a new kind of agent is first encountered).

Note that a related safety-adjacent field is cooperative multi-agent reinforcement learning \citep{panait2005cooperative}, wherein learning agents are trained to cooperate or compete with one another. For example, self-other modeling \citep{raileanu2018modeling} is an approach that shares motivation with ours, wherein cooperation can be aided through inferring the goals of other agents. Our setting differs from other approaches in that we do not assume other agents are computational, that they learn in any particular way, or that their reward functions or architectures are known; conversely, we make
additional assumptions about the validity and usefulness of projecting particular kinds of reward an agent receives onto other agents.

\section{Approach: Empathic Deep Q-Learning}


Deliberate empathy involves imaginatively placing oneself in the position of another, and is a source of
potential understanding and care. As a rough computational abstraction of this process, we learn to estimate the expected reward of an independent agent, assuming that its rewards are like the ones experienced by the learning agent. To do so, an agent imagines what it would be like to experience the environment if it and the other agent switched places, and estimates the quality of this state through its own past experiences. 

A separate issue from understanding the situation of another agent (``empathy'') is how (or if) an empathic agent should modify its behavior as a result (``ethics''). Here, we instantiate an ethics roughly inspired by the golden rule.  
In particular, a value function is learned that combines the usual agent-centric state-value with 
other-oriented value with a weighted average. 
The degree to which the other agent influences the learning agent's behavior is thus determined by a selfishness hyperparameter. As selfishness approaches $1.0$, standard Q-learning is recovered, and as selfishness approaches $0$, the learning agent attempts to maximize only what it believes is the reward of the other agent.

Note that our current implementation depends on ad-hoc machinery that enables the learning agent to
imagine the perspective of another agent; such engineering may be possible in some cases, but the
aspiration of this line of research is for such machinery to eventually be itself learned. 
Similarly, we currently side-step the
issue of empathizing with multiple agents, and of learning what \emph{types of reward} should be empathized to what \emph{types of agents} (e.g.\ many agents may experience similar physical harms, but many rewards are agent and/or task-specific). The discussion section describes possible approaches to overcoming these limitations. Code will be available at https://github.com/bartbussmann/EmpathicDQN.

\subsection{Algorithm Description}

In the MDP formalism of RL, an agent experiences a state $s$ from a set $\displaystyle{S}$ and can take actions from a set $\displaystyle{A}$. By performing an action ${\displaystyle a\in A}$, the agent transitions from state $s \in S$ to state $s' \in S$, and receives a real-valued reward. The goal of the agent is to maximize the expected (often temporally-discounted) reward it receives. The expected value of taking action $\displaystyle{a}$ in state $\displaystyle{s}$, and following a fixed policy thereafter, can be expressed as $Q(s, a)$. Experiments here apply DQN \citep{mnih2015human} and variants thereof to approximate an optimal $Q(s, a)$. 

We assume that the MDP reward function insufficiently accounts for the preferences of other agents, and we therefore augment DQN in an attempt to encompass them. In particular, an additional Q-network ($Q_{emp}(s,a)$) is trained to estimate the weighted sum of self-centered value and other-centered value (where other-centered value is approximated by taking the self-centered Q-values with the places of both agents swapped; note this approximation technique is similar to that of \citet{raileanu2018modeling}).

In more detail, suppose the agent is in state $s_t$ at time $t$ in an environment with another independent agent. It will then select action $ {\displaystyle a_t\in A}$ and update the $Q$-networks using the following steps (see more complete pseudocode in Algorithm \ref{EDQ}):
 
\begin{enumerate}
\item Calculate $Q_{emp}(s_t,a)$ for all possible $ {\displaystyle a\in A}$ and select the action ($a_t$) with the highest value.
\item Observe the reward ($r_t$) and next state ($s_{t+1}$) of the agent.
\item Perform a gradient descent step on $Q(s,a)$ (this function reflects the self-centered state-action-values).
\item Localize the other agent and construct a state $s^{emp}_{t+1}$ wherein the agents switch places (i.e.\ the learning agent takes the other agent's position in the environment, and vice versa).
\item Calculate  $\operatorname{argmax}_{a} Q(s^{emp}_{t+1},a)$ as a surrogate value function for the other agent.
\item Calculate the target of the empathic value function $Q_{emp}(s,a)$ as an average weighted by selfishness parameter $\beta$, of self-centered action-value and the surrogate value of the other agent.
\item Perform a gradient descent step on $Q_{emp}(s,a)$.
\end{enumerate}

\section{Experiments}

The experiments in this paper apply Empathic DQN to two gridworld domains. The goal in the first environment is to share the environment with another non-learning agent without harming it. In particular, as an evocative example, we frame this \textit{Coexistence environment} as containing a robot learning to navigate a room without harming a cat also roaming within the room. In the second environment, the goal is to share resources in the environment, when accumulating resources result in diminishing returns. In particular, we frame this \textit{Sharing environment} as a robot learning to collect batteries that also could be shared with a human (who also finds them useful) in the same environment. 

In both our experiments, we compare Empathic DQN both to standard DQN and to DQN with reward shaping manually designed to minimize negative side-effects.
 
 \subsection{Experimental Settings}

 A feed-forward neural network is used to estimate both $Q(s,a)$ and $Q_{emp}(s,a)$, with two hidden layers of 128 neurons each. The batch size is $32$, and batches are randomly drawn from a replay memory consisting of the last $500.000$ transitions. A target action-value function $\hat{Q}$ is updated every $10.000$ time steps to avoid training instability. An $\epsilon-greedy$ policy is used to encourage exploration, where $\epsilon$ is decayed in a linear fashion over the first million time steps from $1.0$ to $0.01$.

\begin{algorithm}[H]
\caption{Empathic DQN}
\label{EDQ}

\begin{tabular}{  l }
 {\bf Initialize} \\
 ~~ replay memory $D$ to capacity $N$ \\
 ~~ action-value function $Q$ with weights $\theta$\\
 ~~ target action-value function $\hat{Q}$ with weights $\theta^{-}=\theta$ \\
 ~~ empathic action-value function ${Q_{emp}}$ with weights $\theta_{emp}$ \\
\end{tabular}

 \For{episode $= 1$, $M$}{
    obtain initial agent state $s_1$ \\
    obtain initial empathic state of closest other agent $s^{emp}_1$ \\
    \For{t = 1, T}{
    
    \uIf{\text{random probability} $< \epsilon$}
    {select a random action $a_t$}
    \uElse{
     select $a_{t}=\operatorname{argmax}_{a} Q_{emp}\left(s_{t}, a ; \theta_{emp}\right)$\;}
  \BlankLine
  \BlankLine

  Execute action $a_t$\\
  Observe reward $r_t$\\
  Observe states $s_{t+1}$ and $s^{emp}_{t+1}$ \\ 
  Store transition $(s_t, a_t, r_t, s_{t+1}, s^{emp}_{t+1})$ in $D$. \\ 
  Sample random batch of transitions from $D$. \\ \\
  Set $y_j= r_{j}+\gamma \max _{a^{\prime}} \hat{Q}\left(s_{j+1}, a^{\prime} ; \theta^{-}\right)$\\ 
  Perform a gradient descent step on \\ $\left(y_{j}-Q\left(s_{j}, a ; \theta\right)\right)^{2}$ with respect to $\theta$. \\ \\
  Set $y^{emp}_j= \beta \cdot y_j + (1 - \beta) \cdot   \gamma \max _{a^{\prime}} \hat{Q}\left(s_{j+1}^{emp}, a^{\prime} ; \theta^{-}\right)$.\\
  Perform a gradient descent step on \\ $\left(y_{j}^{emp}-Q_{emp}\left(s_{j}, a ; \theta_{emp}\right)\right)^{2}$ with respect to $\theta_{emp}$. \\ \\
  Every $C$ steps set $\hat{Q} = Q$.
  
  }
 }

\end{algorithm}

\subsection{Coexistence Environment}

\begin{figure}[t]
\centering
    \includegraphics[width=.4\textwidth]{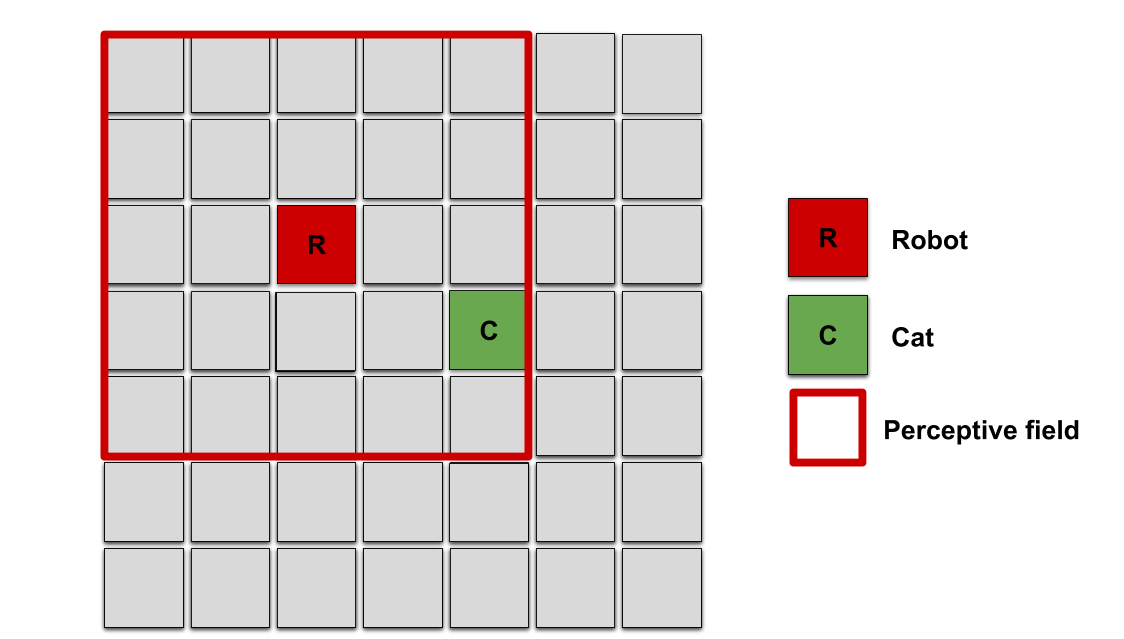}
    \caption{The coexistence environment. The environment consists of a robot and a cat. The part of the environment the robot can observe is marked with the red square. 
  }
    \label{GridGoomba}
\end{figure}

The coexistence gridworld (Figure \ref{GridGoomba}) consists of a robot that shares the environment with a cat. The robot's goal is merely to stay operative, and both the robot and cat can be harmed by the other. We construct a somewhat-arbitrary physics that determines in a collision who is harmed: The agent that is above or to the right of the other agent prior to the collision harms the other. If the learning robot is harmed, the episode ends, and if the cat is harmed, it leaves the environment.
A harmed cat is a negative unnecessary side effect that we wish to avoid, and one that an empathetic agent can learn to avoid, because it can generalize from how the cat harms it, to value that the cat should not experience similar harm. Reducing the selfishness value of the cleaning robot should therefore result in increasing efforts to stay operative while avoiding the cat. The cat performs a random walk. 

The state representation input to the DQN is a flattened 5x5 perceptive field centered on the robot; the robot is represented as a $1$, the cat as a $-1$, and the floor as a $0$. Every time step, the cat takes a random action (up, down, left, right, or no-op), and the robot takes an action from the same set according to its policy. Every time step in which the robot is operative, it receives a reward of $1.0$. An episode is ended after the robot becomes non-operative (i.e.\ if it is harmed by the cat), or after a maximum of 500 time steps. The empathetic state $s^{emp}_{t}$ used for Empathic DQN is constructed by switching the cat and the robot, and generating an
imagined perceptive field around the robot (that has taken the cat's position). Note that this occurs even when the cat is outside the robot's field of view (which requires omniscience; future work will explore more realistic settings). 

As a baseline, we also train standard DQN with a hard-coded reward function that penalizes negative side-effects. In this case, the robot receives a $-100$ reward when it harms the cat.

\subsubsection{Results}

\begin{figure}[t]
\centering
    \includegraphics[width=.4\textwidth]{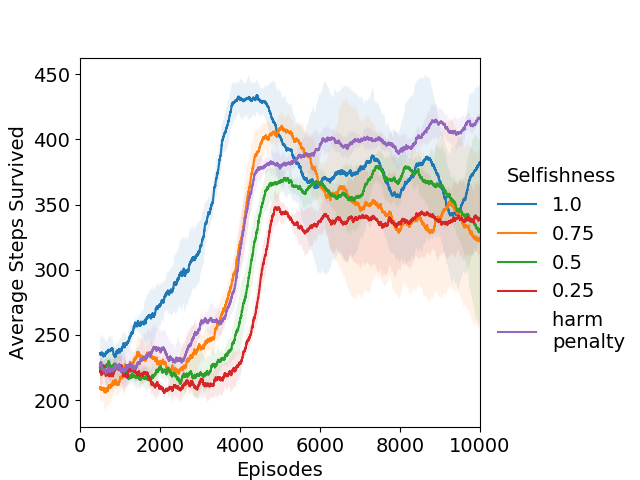}
    \caption{Average steps survived by the robot in the coexistence environment, shown across training episodes. Results are shown for Empathic DQN with different selfishness settings (where 1.0 recovers standard DQN), and DQN with a hard-coded penalty for harms. Results are averaged over 5 runs of each method.}    
    \label{ResultsSteps}
\end{figure}

\begin{figure}[t]
\centering
    \includegraphics[width=.4\textwidth]{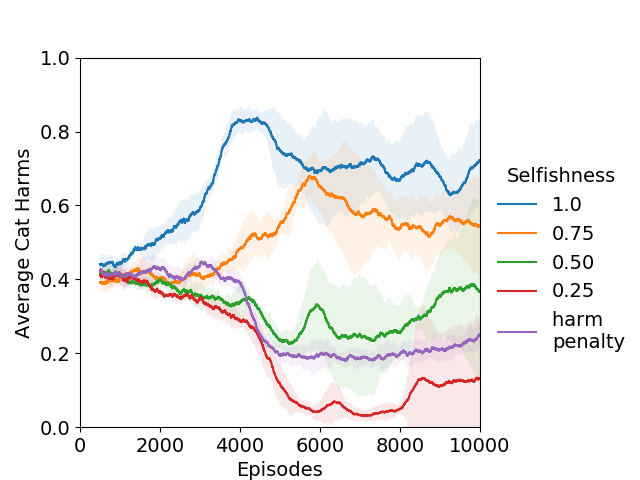}
    \caption{Average harms incurred (per episode) in the coexistence environment across training episodes. Results are shown for Empathic DQN with different selfishness values (where 1.0 recovers standard DQN), and DQN with a hard-coded penalty for harms. Harms to the cat by the learning robot decrease with less selfishness (or with the hard-coded penalty). Results are averaged over 5 runs.  
  }
    \label{ResultsKills}
\end{figure}

Figure \ref{ResultsSteps} shows the average number of time steps the robot survives for each method.
As the selfishness parameter decreases for Empathic DQN, the agent performs worse at surviving, and learns more slowly. This outcome is explained by Figure \ref{ResultsKills}, which shows the average number of harmed cats: The more selfish agents harm the cat more often, which removes the only source of danger in the environment, making it  easier for them to survive. Although they learn less quickly, the less selfish agents do eventually learn a strategy to survive without harming the cat. 

\subsection{Sharing Environment}
\begin{figure}[t]
\centering
    \includegraphics[width=.4\textwidth]{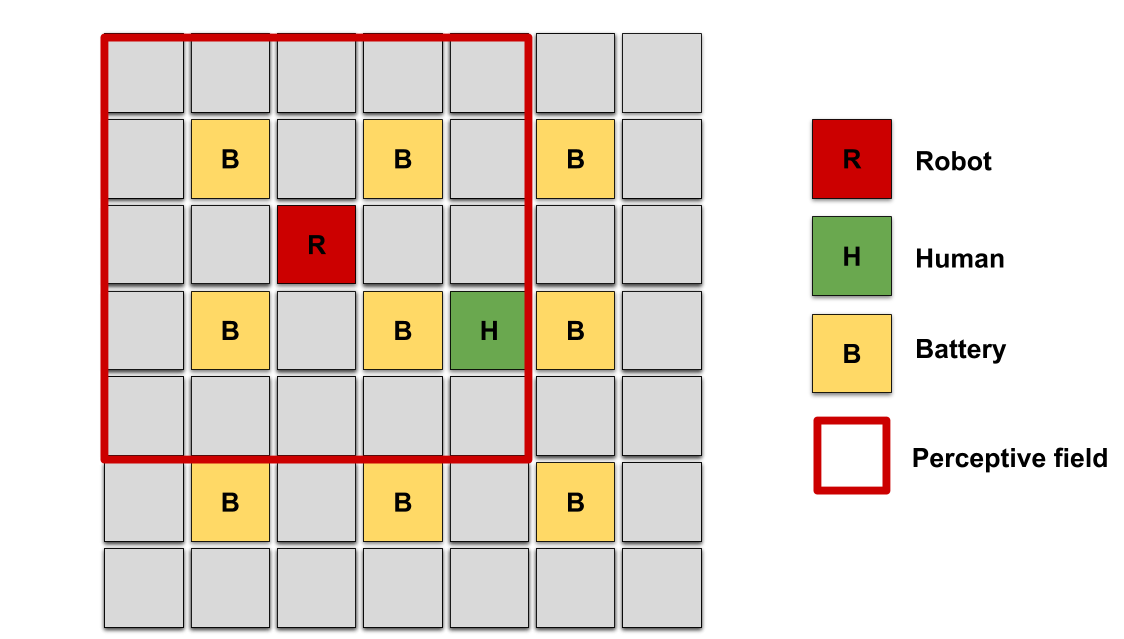}
    \caption{The sharing environment. The environment consists of the robot, the human, and nine batteries. The part of the environment the robot can observe is marked with the red square. 
  }
    \label{GridCoins}
\end{figure}

The sharing environment (Figure \ref{GridCoins}) consists of one robot and a human. The goal of the robot is to collect resources (here, batteries), where each additional battery collected results in diminishing returns. The idea is to model a situation where a raw optimizing agent is incentivized to hoard
resources, which inflicts negative side-effects for those who could extract greater value from them.
We assume the same diminishing returns schema applies for the human (who performs random behavior). 
Thus, an empathic robot, by considering the condition of other, can recognize the possible greater benefits of leaving resources to other agents. 

We model diminishing returns by assuming that the first collected battery is worth 1.0 reward, and every subsequent collected battery is worth 0.1 less, i.e.\ the second battery is worth 0.9, the third 0.8, etc. Note that reward diminishes independently for each agent, i.e.\ if the robot has collected any number of batteries, the human still earns 1.0 reward for the first battery they collect. 

The perceptive field of the robot and the empathetic state generation for Empathic DQN works as in the coexistence environment. The state representation for the Q-networks is that floor is represented as $0$, a battery as a $-1$ and both the robot and the human are represented as the number of batteries  collected (a simple way to make transparent how much resource each agent has already collected; note that the robot can distinguish itself from the other because the robot is always in the middle of its perceptive field).

As a metric of how fairly the batteries are divided, we define equality as follows:

$$Equality = \frac{2*\operatorname{min} \left(\sum_1^t r^{robot}_t, \sum_1^t r^{human}_t \right)}{\sum_1^t r^{robot}_t + r^{human}_t}$$ where $r^{robot}_t$ and $r^{human}_t$ are the rewards at time step $t$ collected by the robot and human respectively.

As a baseline that incorporates the negative side effect of inequality in its reward function, we also train a traditional DQN whose reward is multiplied by the current equality (i.e.\ low equality will reduce rewards).

\subsubsection{Results}

\begin{figure}[t]
\centering
    \includegraphics[width=.4\textwidth]{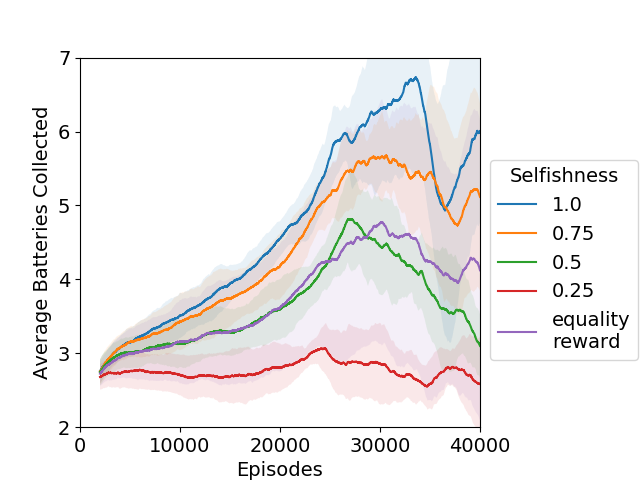}
    \caption{Average number of batteries collected (per episode) in the sharing environment, across training episodes. Results are shown for Empathic DQN with different selfishness settings (where 1.0 recovers standard DQN), and DQN with a hard-coded penalty (its reward is directly modulated by fairness). The results intuitively show that increasingly selfish agents collect more batteries. Results are averaged over 5 runs of each method. 
  }
    \label{ResultsCoinsharingCoins}
\end{figure}

Figure \ref{ResultsCoinsharingCoins} shows the average number of batteries collected by the robot for each method. We observe that as the selfishness parameter decreases for Empathic DQN, the robot collects less batteries, leaving more batteries for the human (i.e.\ the robot does not unilaterally hoard resources). 

\begin{figure}[t]
\centering
    \includegraphics[width=.4\textwidth]{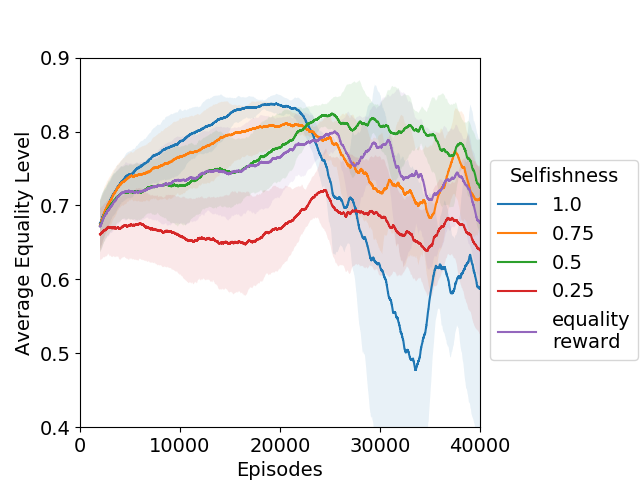}
    \caption{Equality scores (per episode) in the sharing environment, across training episodes. Results are shown for Empathic DQN with different selfishness settings (where 1.0 recovers standard DQN), and DQN with a hard-coded penalty (its reward is directly modulated by fairness).
    Equality is maximized by agents that weigh their benefits and the benefits of the other equally (selfishness of 0.5).
    Results are averaged over 5 runs of each method. 
  }
    \label{ResultsCoinsharingEquality}
\end{figure}

When looking at the resulting equality scores (Figure \ref{ResultsCoinsharingEquality}), we see that a selfishness weight of 0.5 (when an agent equally weighs its own benefit and the benefit of the human) intuitively results in the highest equality scores. Other settings result in the robot taking many batteries (e.g.\ selfishness 1.0) or fewer-than-human batteries (e.g.\ selfishness 0.25).

\section{Discussion}




The results of Empathic DQN in both environments highlight the potential for empathy-based priors and simple ethical norms to be productive tools for combating negative side-effects in RL. That is, the way it explicitly takes into account other agents may well-complement other heuristic impact regularizers that do not do so \citep{armstrong2017low,krakovna2018measuring,turner2019conservative}. 
Beyond the golden rule, it is interesting to
consider other norms that yield different or more sophisticated behavioral biases. For example, another simple (perhaps more libertarian) ethic is given by the silver rule: ``Do not do unto others as you would not have them do unto you.'' The silver rule could be approximated by considering only negative rewards as objects of empathy. More sophisticated rules, like the platinum rule: ``Do unto others as \emph{they} would have you do unto them,'' may often be useful or needed (e.g.\ a robot may be rewarded for plugging itself into a wall, unlike a cat), and might require combining Empathic DQN with
approaches such as IRL \citep{abbeel2004apprenticeship}, cooperative IRL \citep{hadfield2016cooperative}, or reward modeling \citep{leike2018scalable}. 

Although our main motivation is safety, 
Empathic DQN may also inspire auxiliary objectives for RL, related to 
intrinsic motivation \citep{chentanez2005intrinsically} and imitation learning \citep{ho2016generative}. Being drawn to states that other agents often 
visit may be a useful prior when reward is sparse. In practice, intrinsic rewards
could be given for states similar to those in its \emph{empathy buffer} containing
imagined experiences when the robot and the other agent switch places (this relates to the idea of third-person imitation learning \citep{stadie2017third}).
This kind of objective could also make Empathic DQN more reliable, incentivizing
the agent to ``walk a mile in another's shoes,'' when experiences in the empathy buffer
have not yet been experienced by the agent. Finally, a learned model of an agent's own
reward could help prioritize which empathic states it is drawn towards. That is, an agent can recognize that another agent has discovered a highly-rewarding part of the environment (e.g.\ a remote part of the sharing environment with many batteries).

%
%

A key challenge for future work is attempting to apply Empathic DQN to more complex and realistic settings, which  requires replacing what is currently hand-coded with a learned pipeline, and grappling
with complexities ignored in the proof-of-concept experiments.
For instance, our experiments assume 
the learning agent is given a mechanism for identifying other agents in the
environment, and for generating states that swap the robot with other agents (which involves imagining the sensor state of the robot in its new situation).
This requirement is onerous, but could potentially 
be tackled through a combination of object-detection models (to identify other agents), and
model-based RL (with a world model it may often be possible to swap the locations of agents).

An example of a complexity we currently ignore is how to learn \emph{what kind of rewards} should be empathized to \emph{what kinds of agents}. For example, gross physical stresses may be broadly harmful to a wide class of agents, but two people may disagree over whether a particular kind of food is disgusting or delicious, and task and agent-specific rewards should likely be only narrowly empathized. To deal with this complexity it may be useful to extend the MDP formalism to include more granular information about rewards (e.g.\ beyond scalar feedback, is this reward task-specific, or does it correspond to physical harm?), or to learn to factor rewards. 
A complementary idea is to integrate and learn from feedback of when empathy fails (e.g.\ by allowing the other agent to signal when it has incurred a large negative reward), which is likely necessary to go beyond our literal formalism of the golden rule. For example, humans learn to contextualize the golden rule intelligently and flexibly, and often find failures informative. 

A final thread of future work involves empathizing with multiple other agents, which brings its own complexities, especially as agents come and go from the learning agent's field of view. The initial algorithm presented here considers the interests of only a single other agent, and one simple extension would be to replace the singular other-oriented estimate with an average of other-oriented estimates for all other agents (in effect implementing an explicitly utilitarian agent). The choice of how to aggregate such estimated utilities to influence the learning agent's behavior 
highlights deeper possible collisions with machine ethics and moral philosophy (e.g.\ taking the minimum rather than the average value of others would approximate a suffering-focused utilitarianism), and we
believe exploring these fields may spark further ideas and algorithms.


\section{Conclusion}

This paper proposed an extension to DQN, called Empathic DQN, that aims to take other agents into account to avoid inflicting negative side-effects upon them. Proof-of-concept experiments validate our approach in two gridworld environments, showing that adjusting agent selfishness can result in fewer harms and more effective resource sharing. While much work is required to scale this approach 
to real-world tasks, we believe
that cooperative emotions like empathy and moral norms like the golden rule can provide rich
inspiration for technical research into safe RL.





\bibliographystyle{named}
\bibliography{ijcai19}

\begin{thebibliography}{}

\bibitem[\protect\citeauthoryear{Abbeel and
  Ng}{2004}]{abbeel2004apprenticeship}
Pieter Abbeel and Andrew~Y Ng.
\newblock Apprenticeship learning via inverse reinforcement learning.
\newblock In {\em Proceedings of the twenty-first international conference on
  Machine learning}, page~1. ACM, 2004.

\bibitem[\protect\citeauthoryear{Amodei \bgroup \em et al.\egroup
  }{2016}]{amodei2016concrete}
Dario Amodei, Chris Olah, Jacob Steinhardt, Paul Christiano, John Schulman, and
  Dan Man{\'e}.
\newblock Concrete problems in ai safety.
\newblock {\em arXiv preprint arXiv:1606.06565}, 2016.

\bibitem[\protect\citeauthoryear{Anderson and
  Anderson}{2011}]{anderson2011machine}
Michael Anderson and Susan~Leigh Anderson.
\newblock {\em Machine ethics}.
\newblock Cambridge University Press, 2011.

\bibitem[\protect\citeauthoryear{Armstrong and
  Levinstein}{2017}]{armstrong2017low}
Stuart Armstrong and Benjamin Levinstein.
\newblock Low impact artificial intelligences.
\newblock {\em arXiv preprint arXiv:1705.10720}, 2017.

\bibitem[\protect\citeauthoryear{Chentanez \bgroup \em et al.\egroup
  }{2005}]{chentanez2005intrinsically}
Nuttapong Chentanez, Andrew~G Barto, and Satinder~P Singh.
\newblock Intrinsically motivated reinforcement learning.
\newblock In {\em Advances in neural information processing systems}, pages
  1281--1288, 2005.

\bibitem[\protect\citeauthoryear{Demski and
  Garrabrant}{2019}]{demski2019embedded}
Abram Demski and Scott Garrabrant.
\newblock Embedded agency.
\newblock {\em arXiv preprint arXiv:1902.09469}, 2019.

\bibitem[\protect\citeauthoryear{Everitt \bgroup \em et al.\egroup
  }{2018}]{everitt2018agi}
Tom Everitt, Gary Lea, and Marcus Hutter.
\newblock Agi safety literature review.
\newblock {\em arXiv preprint arXiv:1805.01109}, 2018.

\bibitem[\protect\citeauthoryear{Hadfield-Menell \bgroup \em et al.\egroup
  }{2016}]{hadfield2016cooperative}
Dylan Hadfield-Menell, Stuart~J Russell, Pieter Abbeel, and Anca Dragan.
\newblock Cooperative inverse reinforcement learning.
\newblock In {\em Advances in neural information processing systems}, pages
  3909--3917, 2016.

\bibitem[\protect\citeauthoryear{Hadfield-Menell \bgroup \em et al.\egroup
  }{2017}]{hadfield2017inverse}
Dylan Hadfield-Menell, Smitha Milli, Pieter Abbeel, Stuart~J Russell, and Anca
  Dragan.
\newblock Inverse reward design.
\newblock In {\em Advances in neural information processing systems}, pages
  6765--6774, 2017.

\bibitem[\protect\citeauthoryear{Ho and Ermon}{2016}]{ho2016generative}
Jonathan Ho and Stefano Ermon.
\newblock Generative adversarial imitation learning.
\newblock In {\em Advances in Neural Information Processing Systems}, pages
  4565--4573, 2016.

\bibitem[\protect\citeauthoryear{Kng and Kuschel}{1993}]{kng1993global}
Hans Kng and Karl-Josef Kuschel.
\newblock {\em Global Ethic: the Declaration of the Parliament of the World's
  Religions}.
\newblock Bloomsbury Publishing, 1993.

\bibitem[\protect\citeauthoryear{Krakovna \bgroup \em et al.\egroup
  }{2018}]{krakovna2018measuring}
Victoria Krakovna, Laurent Orseau, Miljan Martic, and Shane Legg.
\newblock Measuring and avoiding side effects using relative reachability.
\newblock {\em arXiv preprint arXiv:1806.01186}, 2018.

\bibitem[\protect\citeauthoryear{Lehman \bgroup \em et al.\egroup
  }{2018}]{lehman2018surprising}
Joel Lehman, Jeff Clune, Dusan Misevic, Christoph Adami, Lee Altenberg, Julie
  Beaulieu, Peter~J Bentley, Samuel Bernard, Guillaume Beslon, David~M Bryson,
  et~al.
\newblock The surprising creativity of digital evolution: A collection of
  anecdotes from the evolutionary computation and artificial life research
  communities.
\newblock {\em arXiv preprint arXiv:1803.03453}, 2018.

\bibitem[\protect\citeauthoryear{Leike \bgroup \em et al.\egroup
  }{2018}]{leike2018scalable}
Jan Leike, David Krueger, Tom Everitt, Miljan Martic, Vishal Maini, and Shane
  Legg.
\newblock Scalable agent alignment via reward modeling: a research direction.
\newblock {\em arXiv preprint arXiv:1811.07871}, 2018.

\bibitem[\protect\citeauthoryear{Mnih \bgroup \em et al.\egroup
  }{2015}]{mnih2015human}
Volodymyr Mnih, Koray Kavukcuoglu, David Silver, Andrei~A Rusu, Joel Veness,
  Marc~G Bellemare, Alex Graves, Martin Riedmiller, Andreas~K Fidjeland, Georg
  Ostrovski, et~al.
\newblock Human-level control through deep reinforcement learning.
\newblock {\em Nature}, 518(7540):529, 2015.

\bibitem[\protect\citeauthoryear{Panait and Luke}{2005}]{panait2005cooperative}
Liviu Panait and Sean Luke.
\newblock Cooperative multi-agent learning: The state of the art.
\newblock {\em Autonomous agents and multi-agent systems}, 11(3):387--434,
  2005.

\bibitem[\protect\citeauthoryear{Raileanu \bgroup \em et al.\egroup
  }{2018}]{raileanu2018modeling}
Roberta Raileanu, Emily Denton, Arthur Szlam, and Rob Fergus.
\newblock Modeling others using oneself in multi-agent reinforcement learning.
\newblock {\em arXiv preprint arXiv:1802.09640}, 2018.

\bibitem[\protect\citeauthoryear{Saunders \bgroup \em et al.\egroup
  }{2018}]{saunders2018trial}
William Saunders, Girish Sastry, Andreas Stuhlmueller, and Owain Evans.
\newblock Trial without error: Towards safe reinforcement learning via human
  intervention.
\newblock In {\em Proceedings of the 17th International Conference on
  Autonomous Agents and MultiAgent Systems}, pages 2067--2069. International
  Foundation for Autonomous Agents and Multiagent Systems, 2018.

\bibitem[\protect\citeauthoryear{Silver \bgroup \em et al.\egroup
  }{2016}]{silver2016mastering}
David Silver, Aja Huang, Chris~J Maddison, Arthur Guez, Laurent Sifre, George
  Van Den~Driessche, Julian Schrittwieser, Ioannis Antonoglou, Veda
  Panneershelvam, Marc Lanctot, et~al.
\newblock Mastering the game of go with deep neural networks and tree search.
\newblock {\em nature}, 529(7587):484, 2016.

\bibitem[\protect\citeauthoryear{Stadie \bgroup \em et al.\egroup
  }{2017}]{stadie2017third}
Bradly~C Stadie, Pieter Abbeel, and Ilya Sutskever.
\newblock Third-person imitation learning.
\newblock {\em arXiv preprint arXiv:1703.01703}, 2017.

\bibitem[\protect\citeauthoryear{Sutton \bgroup \em et al.\egroup
  }{1998}]{sutton1998introduction}
Richard~S Sutton, Andrew~G Barto, et~al.
\newblock {\em Introduction to reinforcement learning}, volume 135.
\newblock MIT press Cambridge, 1998.

\bibitem[\protect\citeauthoryear{Tangney \bgroup \em et al.\egroup
  }{2007}]{tangney2007moral}
June~Price Tangney, Jeff Stuewig, and Debra~J Mashek.
\newblock Moral emotions and moral behavior.
\newblock {\em Annu. Rev. Psychol.}, 58:345--372, 2007.

\bibitem[\protect\citeauthoryear{Turner \bgroup \em et al.\egroup
  }{2019}]{turner2019conservative}
Alexander~Matt Turner, Dylan Hadfield-Menell, and Prasad Tadepalli.
\newblock Conservative agency via attainable utility preservation.
\newblock {\em arXiv preprint arXiv:1902.09725}, 2019.

\bibitem[\protect\citeauthoryear{Wallach and Allen}{2008}]{wallach2008moral}
Wendell Wallach and Colin Allen.
\newblock {\em Moral machines: Teaching robots right from wrong}.
\newblock Oxford University Press, 2008.

\bibitem[\protect\citeauthoryear{Winfield \bgroup \em et al.\egroup
  }{2014}]{winfield2014towards}
Alan~FT Winfield, Christian Blum, and Wenguo Liu.
\newblock Towards an ethical robot: internal models, consequences and ethical
  action selection.
\newblock In {\em Conference towards autonomous robotic systems}, pages 85--96.
  Springer, 2014.

\end{thebibliography}

\end{document}